\documentclass[sigconf, authorversion, nonacm]{acmart}

\usepackage{graphicx}
\usepackage{hyperref}

\usepackage{amsmath}

\usepackage[]{mdframed}

\usepackage{listings}
\usepackage{xcolor}
\usepackage[export]{adjustbox}
\usepackage{subfig}
\usepackage{booktabs}
\usepackage{color}
\usepackage{tabularray}
\usepackage[nodayofweek]{datetime}

\definecolor{codegreen}{rgb}{0,0.6,0}
\definecolor{codegray}{rgb}{0.5,0.5,0.5}
\definecolor{codepurple}{rgb}{0.58,0,0.82}
\definecolor{backcolour}{rgb}{0.95,0.95,0.92}

\lstdefinestyle{mystyle}{
    backgroundcolor=\color{backcolour},   
    commentstyle=\color{codegreen},
    keywordstyle=\color{magenta},
    numberstyle=\tiny\color{codegray},
    stringstyle=\color{codepurple},
    basicstyle=\ttfamily\scriptsize,   
    breakatwhitespace=false,         
    breaklines=true,                 
    captionpos=b,                    
    keepspaces=true,                 
    numbers=left,                    
    numbersep=5pt,                  
    showspaces=false,                
    showstringspaces=false,
    showtabs=false,                  
    tabsize=2
}
\AtBeginDocument{%
  }

\begin{document}

\title{Empowering Federated Learning for Massive Models with NVIDIA FLARE}
%

%
%
\author{Holger R. Roth, Ziyue Xu, Yuan-Ting Hsieh, Adithya Renduchintala, Isaac Yang, Zhihong Zhang, Yuhong Wen, Sean Yang, Kevin Lu, Kristopher Kersten, Camir Ricketts, Daguang Xu, Chester Chen, Yan Cheng, Andrew Feng}
%
%
%
\affiliation{
  \institution{NVIDIA Corporation}
  \city{Santa Clara}
  \country{USA}
}

\begin{abstract}
In the ever-evolving landscape of artificial intelligence (AI) and large language models (LLMs), handling and leveraging data effectively has become a critical challenge. Most state-of-the-art machine learning algorithms are data-centric. However, as the lifeblood of model performance, necessary data cannot always be centralized due to various factors such as privacy, regulation, geopolitics, copyright issues, and the sheer effort required to move vast datasets. In this paper, we explore how federated learning enabled by NVIDIA FLARE can address these challenges with easy and scalable integration capabilities, enabling parameter-efficient and full supervised fine-tuning of LLMs for natural language processing and biopharmaceutical applications to enhance their accuracy and robustness.
\end{abstract}


\maketitle
\renewcommand{\shortauthors}{Roth et al.}
%
%
\section{Introduction}
Data management and utilization are pivotal challenges in the dynamic realm of artificial intelligence (AI) and large language models (LLMs). Contemporary machine learning algorithms often rely on data-centric approaches and face obstacles in centralized data handling due to multifaceted concerns such as privacy, regulatory constraints, geopolitical factors, copyright issues, and the considerable logistical demands associated with moving extensive datasets. 

This paper delves into the practical application of federated learning (FL), particularly exploring the capabilities offered by NVIDIA FLARE\footnote{\url{https://github.com/NVIDIA/NVFlare}} (NVFlare) ~\cite{roth2022nvidia} in addressing these challenges. Through seamless and scalable integration, NVFlare facilitates parameter-efficient fine-tuning (PEFT)~\cite{he2021towards,ding2023parameter} and full supervised fine-tuning (SFT). With a specific focus on applications in natural language processing (NLP) and biopharma utilizing modern LLM architectures. The aim is to enhance the accuracy and robustness of these models by applying FL in real-world situations.

\paragraph{The Data Challenge}
The need to access data from multiple sources is a common scenario in many LLM tasks. Consider scenarios like gathering reports from different hospitals for medical research or collecting financial data from diverse institutions for analysis. Centralizing such data may be impractical and hindered by privacy concerns, regulatory hurdles, etc. FL offers an elegant solution to this issue~\cite{rieke2020future}.

\paragraph{Federated Learning}
FL has emerged as a practical solution to tackle data challenges. Instead of centrally training models with access to raw data, FL facilitates sharing model updates rather than raw data itself. This means that participating clients can train models locally using their own private datasets to compute a local model update. These local updates are then combined globally to update the model parameters. This approach maintains the privacy of individual datasets while enabling the global model to benefit from the collective knowledge gained during training. The result is the training of more robust and generalizable models~\cite{dayan2021federated}.

FL provides several options for training AI models. Essentially, it allows for the training of a global model while ensuring the privacy and governance of the data involved. Moreover, FL can be tailored to meet the specific needs of each client, thus enabling the creation of personalized models. Additionally, FL infrastructure extends beyond training and can also be utilized for tasks such as inference and federated evaluation.

\section{Methods}
\subsection{FL Framework}
NVFlare is an open-source framework that allows researchers and data scientists to seamlessly move their machine learning and deep learning workflows into a federated paradigm. Furthermore, it empowers platform developers to construct secure and privacy-preserving solutions for collaborative multiparty distributed workflows.
NVFlare is a lightweight, flexible, and scalable FL framework implemented in Python. Notably, it remains agnostic to the underlying training library, allowing developers to use PyTorch, TensorFlow, or even pure NumPy for their data science workflows in a federated setting.
In the NVFlare ecosystem, a standard FL workflow, like the well-known federated averaging (FedAvg) algorithm~\cite{mcmahan2017communication}, involves the following key steps. Each FL client receives an initial global model from the FL server and conducts local training on their data. Next, the clients transmit model updates to the server for aggregation. The server, in turn, applies these aggregated updates to refine the global model for subsequent training rounds. This procedure is repeated until convergence is achieved.
While NVFlare finds frequent use in federated deep learning~\cite{dayan2021federated,roth2020federated,sarma2021federated,guo2022auto,sun2023communication,wang2023condistfl,jiang2023fair,xu2022closing}, its versatility extends to supporting general federated computing across diverse clients. It provides the \textit{Controller Programming API}, enabling researchers to craft flexible workflows for orchestrating client collaboration. FedAvg~\cite{mcmahan2017communication} and cyclic weight transfer~\cite{chang2018distributed} are examples of such workflows.

At the heart of NVFlare lies the concept of collaboration through "tasks." An FL controller assigns tasks (e.g., deep-learning training with model weights) to one or more FL clients, processes returned results (e.g., model weight updates), and may assign additional tasks based on these results and other factors (e.g., a pre-configured number of training rounds). This task-based interaction repeats until the experiment's objectives are met.

\subsection{Easy Adaptation of ML Workflows via Client API}
The NVFlare \textit{Client API} offers a convenient solution for users looking to transition their centralized, local training code to FL with several advantages:

\begin{itemize}
    \item \textbf{Minimal Code Changes:} Users can achieve the transition with only a few lines of code adjustments, eliminating the need for a comprehensive restructuring or implementation of a new class.

    \item \textbf{Simplicity:} The \textit{Client API} minimizes the introduction of new NVFlare-specific concepts to users, streamlining the adaptation process by leveraging familiar programming constructs.

    \item \textbf{Flexibility:} Users can easily adapt existing local training code written in various frameworks such as PyTorch, PyTorch Lightning, and HuggingFace, making the transition seamless and efficient.
\end{itemize}

\noindent The general structure of a popular FL workflow, such as \textit{FedAvg} is as follows:

\begin{mdframed}
    \begin{enumerate}
        \item FL server initializes an initial model.
    
        \item \textit{For each round (global iteration):}
        \begin{enumerate}
            \item FL server sends the global model to clients.
        
            \item Each FL client starts with this global model and trains on their own data.
        
            \item Each FL client sends back their model update.
        
            \item FL server aggregates all the updates and produces a new global model.
        \end{enumerate}
    \end{enumerate}
\end{mdframed}

\noindent On the client side, the training workflow is as follows:

\begin{mdframed}
    \begin{enumerate}
        \item Receive the model from the FL server.
    
        \item Perform local training on the received global model and/or evaluate the received global model for model selection.
    
        \item Send the new model back to the FL server.
    \end{enumerate}
\end{mdframed}

\noindent To convert a centralized training code to FL, we need to adapt the code to execute the following steps:

\begin{mdframed}
    \begin{enumerate}
        \item Obtain the required information from the received model.
    
        \item Run local training.
    
        \item Put the results in a new model to be sent back to the FL server.
    \end{enumerate}
\end{mdframed}

\noindent For a general use case, there are three essential methods for the \textit{Client API}:

\begin{itemize}
    \item \texttt{init()}: Initializes NVFlare Client API environment.
    \item \texttt{receive()}: Receives model from the FL server.
    \item \texttt{send()}: Sends the model to the FL server.
\end{itemize}

\noindent With these simple methods, the developers can use the \textit{Client API} to change their centralized training code to an FL scenario with five lines of code changes as shown in Listing~\ref{lst:client_api}.

\begin{lstlisting}[language=Python, caption=Client API example., label=lst:client_api]
    import nvflare.client as flare
    
    flare.init() # 1. Initializes NVFlare Client API environment.
    input_model = flare.receive() # 2. Receives model from the FL server.
    params = input_model.params # 3. Obtain the required information from the received model.
    
    # original local training code
    new_params = local_train(params)
    
    output_model = flare.FLModel(params=new_params) # 4. Put the results in a new `FLModel`
    flare.send(output_model) # 5. Sends the model to the FL server.    
\end{lstlisting}

\noindent If using standardized training frameworks such as PyTorch Lightning, the conversion to FL can be even more streamlined. As an example in this paper, we use the GPT model from NVIDIA NeMo framework\footnote{\url{https://www.nvidia.com/en-us/ai-data-science/generative-ai/nemo-framework}}~\cite{Harper_NeMo_a_toolkit} to show the application of PEFT and SFT for NLP tasks. NeMo leverages PyTorch Lightning for model training. One notable feature of NVFlare is the \textit{Lightning Client API}, which significantly simplifies the process of converting local training scripts to run in FL scenarios. With just a few lines of code changes, one can seamlessly integrate methods like PEFT and SFT. As shown in Listing~\ref{lst:lightning_client}, the \textit{Lightning trainer} can be adapted to run FL just by calling \texttt{flare.patch(trainer)}. 
Next, an extra while loop (\texttt{while flare.is\_running:}) is added to allow reusing the same trainer object each round of FL.
Optionally, we call \texttt{trainer.validate(model)} to evaluate the global model received from the FL server at the current round on the client’s data. This is useful for enabling global model selection on the server based on validation scores received from each client.

\begin{lstlisting}[language=Python, caption=Pythorch Lightning Client API example., label=lst:lightning_client]
    import nvflare.client.lightning as flare

    ...
    # 1. flare patch
    flare.patch(trainer)

    # 2. Add while loop to keep receiving the model in each FL round.
    # Note, after `flare.patch` the trainer.fit/validate will get the
    # global model internally at each round.
    while flare.is_running():
        # (optional): get the FL system info
        fl_sys_info = flare.system_info()
        print("--- fl_sys_info ---")
        print(fl_sys_info)            

        # 3. evaluate the current global model to allow server-side model selection.
        print("--- validate global model ---")
        trainer.validate(model)

        # 4. Perform local training starting with the received global model.
        print("--- train new model ---")      
        trainer.fit(model)
\end{lstlisting}

\subsection{Server Workflow Implementation}
\begin{figure}[htbp]
    \centering
    \includegraphics[width=0.95\columnwidth]{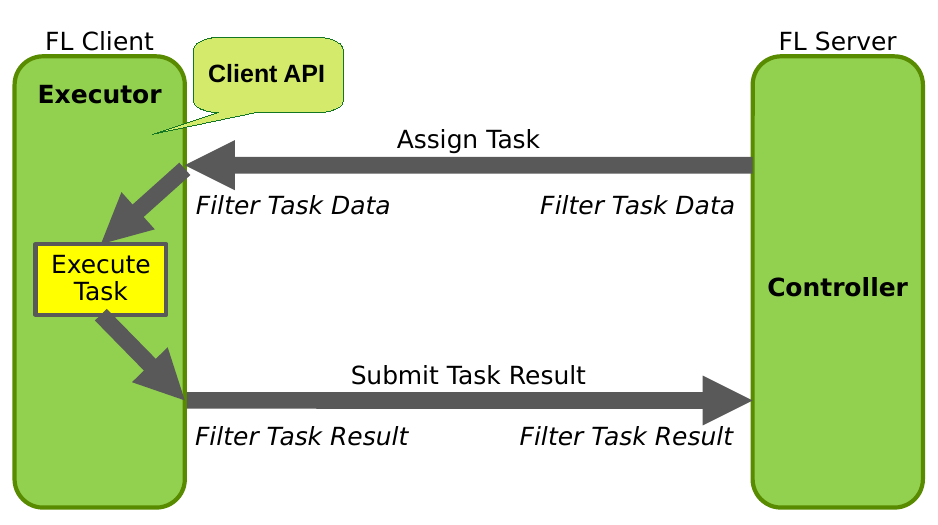}
    \caption{Server workflow \textit{Controller} and \textit{Executor} with \textit{Client API}. \label{fig:controller}}
\end{figure}

NVFlare’s collaborative computing is achieved through the \textit{Controller}/\textit{Executor} interactions. The diagram in Fig.~\ref{fig:controller} shows how the \textit{Controller} and \textit{Executor} interact.
The \textit{Controller} is a class that controls or coordinates the \textit{Executors} to get a job done. It is run on the FL server (highlighted on the right).
A \textit{Executor} is capable of performing tasks. \textit{Executors} run on FL clients and execute the client API described above. In its control logic (it's \texttt{run()} routine), the \textit{Controller} assigns tasks to \textit{Executors} and processes task results from the \textit{Executors}. This allows the easy integration of additional data filters (for example, for adding homomorphic encryption~\cite{zhang2020batchcrypt} or differential privacy filters~\cite{li2019privacy} to the task data or results received or produced by the server or clients).

In Listing~\ref{lst:controller}, we show a simplified implementation of the FedAvg~\cite{mcmahan2017communication} algorithm with NVFlare\footnote{Scheduled for upcoming 2.5.0 release of NVFlare.}. The \texttt{run()} routine implements the main algorithmic logic. Subroutines, like \texttt{sample\_clients()} and \texttt{scatter\_and\_gather\_model()} utilize the \textit{communicator} object, native to each \textit{Controller} to get the list of available clients, distribute the current global model to the clients, and collect their results. For simplicity, we do not show the implementation of the aggregation, model update, and saving routines.
\begin{lstlisting}[language=Python, caption=Federated averaging workflow controller example., label=lst:controller]

class FedAvg(Controller):
   def __init__(self,
                min_clients: int,
                num_rounds: int
                ):
        self.model = ... # initialize the global model
        ...
        
    def run(self) -> None:
        self.info("Start FedAvg.")

        for _current_round in range(self._num_rounds):
            self.info(f"Round {self._current_round} started.")
            # 1. sample the available clients
            clients = self.sample_clients(self._min_clients)
            # 2. send the current global model to clients and receive the model updates
            results = self.scatter_and_gather_model(targets=clients)
            # 3. aggregate the results
            aggregate_results = self.aggregate(results)
            # 4. update the current global model
            self.update_model(aggregate_results)
            # 5. save the current global model
            self.save_model()
            
        self.info("Finished FedAvg.")

    def sample_clients(min_clients):
        # add optional random sampling strategy
        return self.communicator.get_clients()[0:min_clients]

    def scatter_and_gather_model(tagets, data):
        return self.communicator.broadcast_and_wait(
            task_name="train",  
            min_responses=self.min_clients, 
            data=self.model, 
            targets=targets, 
            callback=None)
    ...
\end{lstlisting}
Due to the separation of the controller logic and the communication object (\texttt{self.communicator)}, it is possible to run an NVFlare \textit{Controller} both on the server and the clients, allowing a straightforward implementation of alternative communication strategies such as split learning~\cite{gupta2018distributed} or swarm learning~\cite{warnat2021swarm}.

\subsection{Scalable Model Training via Streaming}
\label{sec:streaming}

\begin{figure*}[htbp]
    \centering
    \includegraphics[width=0.95\textwidth]{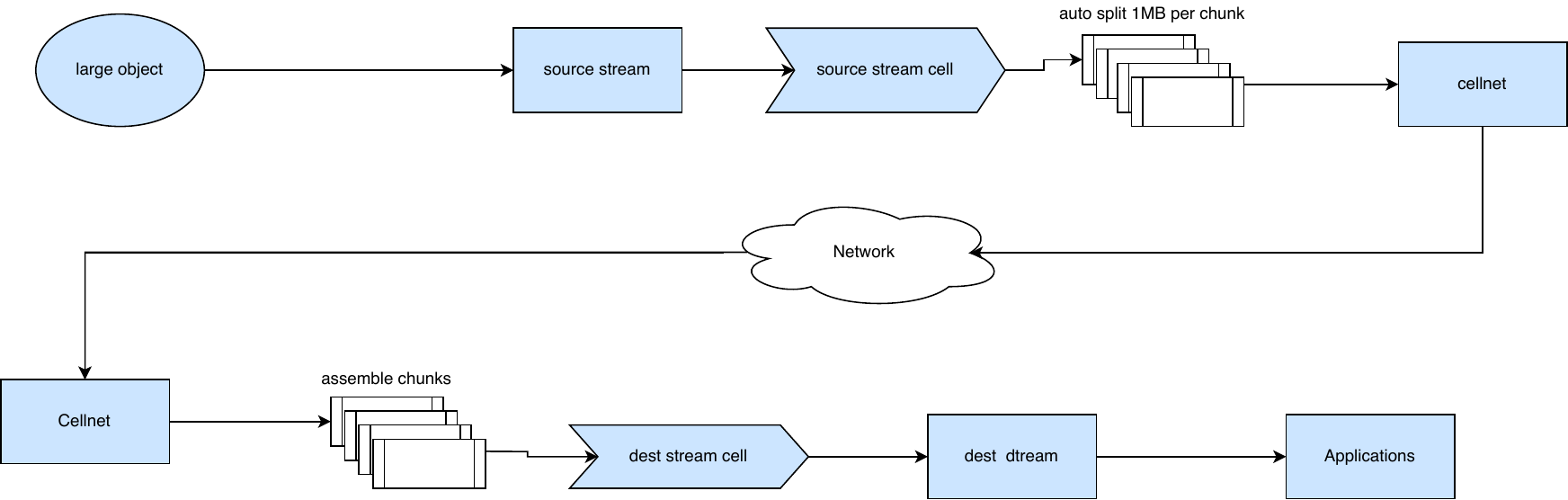}
    \caption{Data streaming API. \label{fig:streaming_api:chunking}}
\end{figure*}

As FL or AI tasks in general become more and more complex, their model sizes increase~\cite{smith2022using}. The size of mainstream LLMs can be enormous, ranging from a few billion parameters to tens of billions of parameters, which leads to a significant increase in model sizes that need to be communicated during FL training. However, using native communication protocols directly can introduce inefficiencies and instability issues. Furthermore, protocols such as gRPC have hard size limits (2 GB) for single messages. Typical model sizes of modern LLMs exceed those limits and can even reach hundreds of GB. NVFlare supports large models as long as the system memory of servers and clients can handle it. However, it requires special considerations because the network bandwidth and, thus, the time to transmit such a large amount of data during an NVFlare job runtime varies significantly. 

In the NVFlare 2.4.0 release, communication capabilities have been significantly enhanced through our new \textit{data streaming API}. 
Our streaming API has four different variations: byte streaming, blob streaming, file streaming, and object streaming, which can work together with different communication protocols (drivers gRPC, HTTP, TCP, etc.). The ``Streamable Framed Message'' (SFM) layer manages the drivers and connections and sends messages. One can change the driver without affecting the upper-layer applications. In other words, one can switch between gRPC, TCP, HTTP, etc., and the applications built on top will work without any changes. One can even build customer drivers that suit their needs. 

As illustrated in Fig.~\ref{fig:streaming_api:chunking}, the large model is now divided into 1 megabyte (MB) chunks and streamed to the target (server or client), bringing a complete transformation to the overall system with the introduction of a new streaming layer designed to handle large data transfers. Once the message arrives at the target end-point, the object is re-assembled to restore the original message payload. Refer to Section~\ref{sec:streaming_results} for quantitative results on large data streaming.


\section{Applications}

\subsection{Adaption of Foundational LLMs}

Foundational LLMs are pre-trained on a vast amount of general text data~\cite{brown2020language}. However, they may not be specialized for specific domains or downstream tasks. Further fine-tuning allows these models to adapt and specialize for particular domains and tasks, making them more effective and accurate in delivering domain- and task-specific results, which is essential to harness their potential and adapt them to various applications' diverse and evolving needs. 

PEFT and SFT are two vital approaches that aim to tailor foundational LLMs to specific domains and tasks efficiently and effectively. Both targets achieve domain and task-specific adaptation based on foundational LLMs. SFT fine-tunes all LLM parameters, while PEFT tries to add adaptation parameters/layers while keeping the LLM parameters fixed, making it a cost-effective and resource-efficient option. These techniques play a pivotal role in harnessing the power of LLMs for a multitude of applications, offering tailored and resource-aware solutions for a wide range of applications.
\begin{figure*}[htbp]
    \centering
     \subfloat[\centering PEFT]{\includegraphics[height=0.35\textwidth,valign=m]{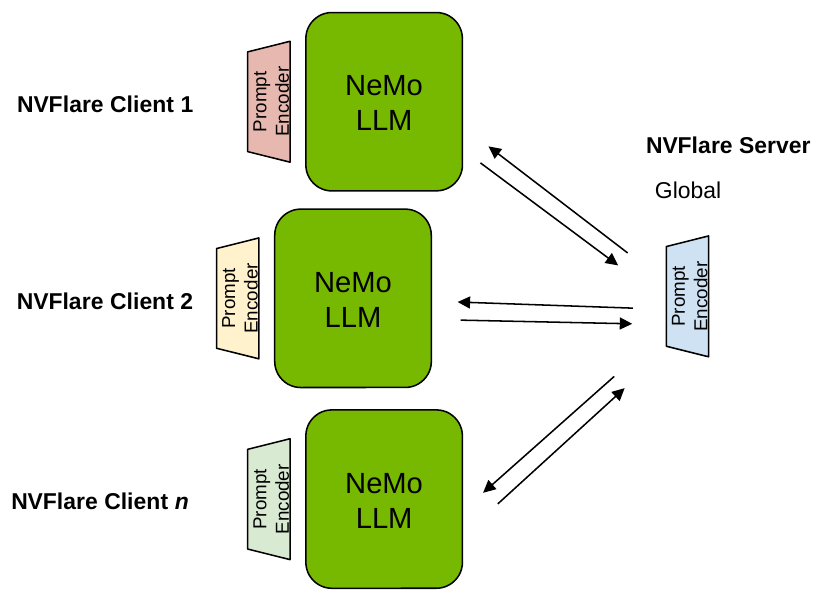}}
     \hfill
     \subfloat[\centering SFT]{\includegraphics[height=0.35\textwidth,valign=m]{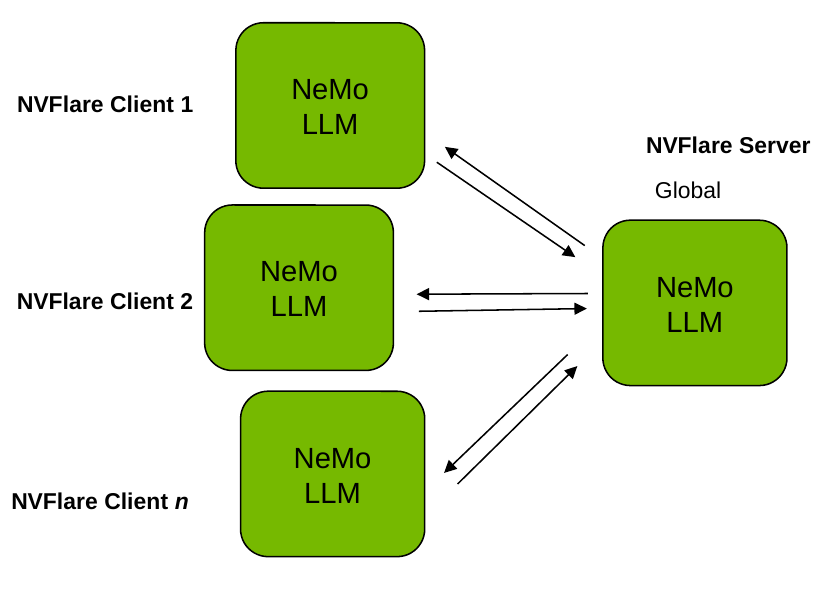}} 
    \caption{Federated parameter-efficient fine-tuning (PEFT) and full supervised fine-tuning (SFT) with global model and $n$ clients. \label{fig:peft_vs_sft}}
\end{figure*}

\subsection{FL for LLM Adaptations}

As with other AI techniques, the performance of LLMs benefits from larger and more diverse datasets. More data usually translates to better accuracy, improved robustness, and generalizability.

As shown in Fig.~\ref{fig:peft_vs_sft}, using PEFT, the parameters of the foundational LLMs are frozen and remain fixed during training and evaluation, while additional parameters are injected for customization. Hence, only these injected parameters are tuned at local clients and aggregated globally. Using SFT, on the other hand, all the parameters of the LLMs are fine-tuned and communicated for aggregation.

As SFT fine-tunes the entire network, the whole model must be transferred and aggregated. This transmission challenge must be properly addressed to enable SFT with recent LLMs in FL using NVFlare's data streaming API (see Section~\ref{sec:streaming}). 

\subsection{Federated Protein Embeddings and Task Model Fitting}
Next, we explore obtaining protein-learned representations in the form of embeddings using an ESM-style pre-trained model~\cite{meier2021language}. The model is trained with NVIDIA's BioNeMo framework\footnote{\url{https://www.nvidia.com/en-us/clara/bionemo}} for LLM training and inference.

Using BioNeMo, users can obtain numerical vector representations of protein sequences called embeddings. Protein embeddings can then be used for visualization or making downstream predictions.

Here, we are interested in training a neural network to predict subcellular location from an embedding.

\paragraph{Subcellular Location Prediction}

\begin{figure}[htbp]
    \centering
    \includegraphics[width=0.95\columnwidth]{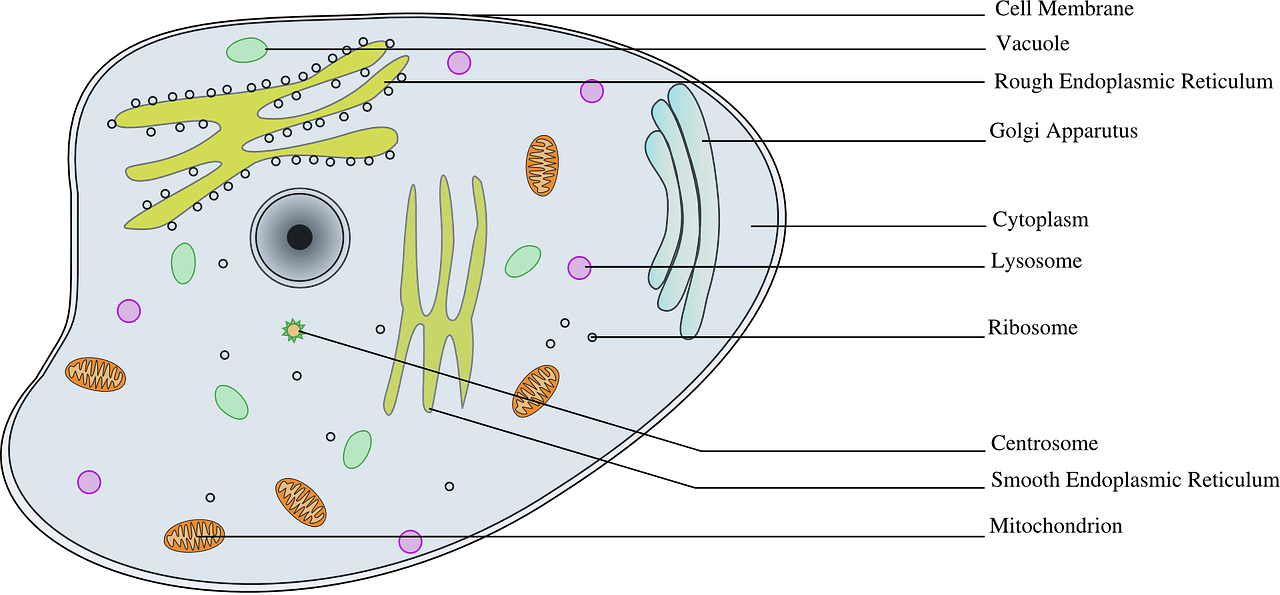}
    \caption{Cross section of an animal cell~\cite{cellimage}.  \label{fig:cell}}
\end{figure}

The data we will be using comes from the work by Stärk et al.~\cite{stark2021light}. In this paper, the authors developed a machine learning algorithm to predict the subcellular location of proteins from sequence through protein language models that are similar to those hosted by BioNeMo. Protein subcellular location refers to where the protein localizes in the cell; for example, a protein may be expressed in the `Nucleus' or the `Cytoplasm.' Knowing where proteins localize can provide insights into the underlying mechanisms of cellular processes and help identify potential targets for drug development. Figure~\ref{fig:cell} includes a few examples of subcellular locations in an animal cell.

We will utilize FASTA\footnote{\url{https://en.wikipedia.org/wiki/FASTA_format}} sequences for our target input sequences in a benchmark dataset called ``Fitness Landscape Inference for Proteins'' (FLIP)~\cite{dallago2021flip}. FLIP encompasses experimental data across adeno-associated virus stability for gene therapy, protein domain B1 stability and immunoglobulin binding, and thermostability from multiple protein families.

\paragraph{Model Architecure} 
We utilize the \textit{ESM-1nv} model developed using the BioNeMo framework. The model uses an architecture called ``Bidirectional Encoder Representations from Transformers'' (BERT) and is based on the ESM-1 model~\cite{rives2021biological,devlin2018bert}. Pre-norm layer normalization and GELU~\cite{hendrycks2016gaussian} activation are used throughout. The model has six layers, 12 attention heads, a hidden space dimension of 768, and contains 44M parameters. The input sequence length is limited to 512 amino acids.

In Section~\ref{sec:scl_results}, we show results for running federated inference to extract protein embeddings from the clients' local data, followed by the application of FedAvg to training an MLP for the downstream subcellular location prediction task.

\section{Results}
\subsection{Large message streaming}
\label{sec:streaming_results}
We have extensively tested the streaming feature across regions and cloud providers, including AWS and Azure, to test the transfer of large model sizes. This example used a randomly initialized model consisting of a dictionary of 64 keys. Each key held a 2GB floating-point number array (resulting in a total model size of 128GB). This message size is much larger than commonly used modern LLMs, such as the \textit{LLama-2} 7B parameter model\footnote{\url{https://huggingface.co/meta-llama/Llama-2-7b}}~\cite{touvron2023llama}, which requires checkpoint sizes of 13.5GB to store all its parameters. Even large models such as \textit{CodeLlama} 70B parameter model\footnote{\url{https://huggingface.co/codellama/CodeLlama-70b-hf}}~\cite{roziere2023code}, which takes $\sim$140GB to store their checkpoints, are similar to our experiment.

\begin{figure*}[htbp]
    \centering
     \subfloat[\centering Server]{\includegraphics[width=0.33\textwidth,valign=m]{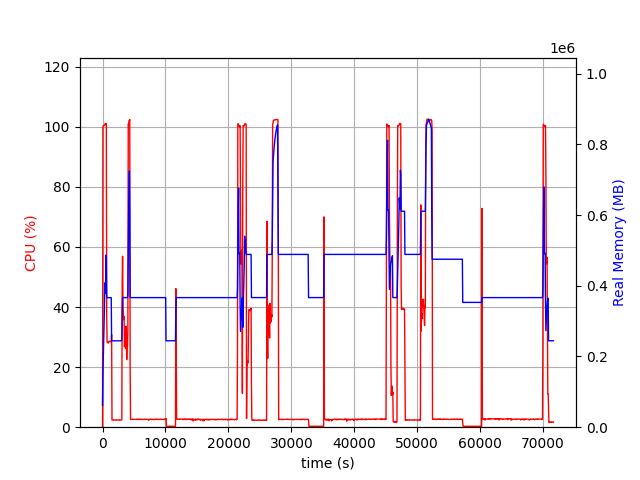}}
     \hfill
     \subfloat[\centering Site-1]{\includegraphics[width=0.33\textwidth,valign=m]{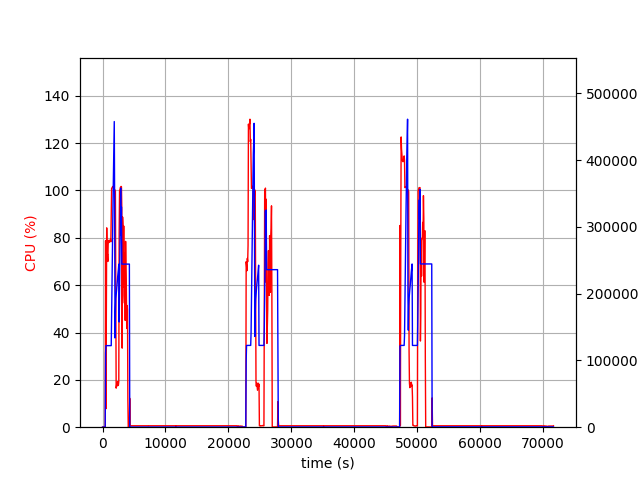}}
     \hfill
     \subfloat[\centering Site-2]{\includegraphics[width=0.33\textwidth,valign=m]{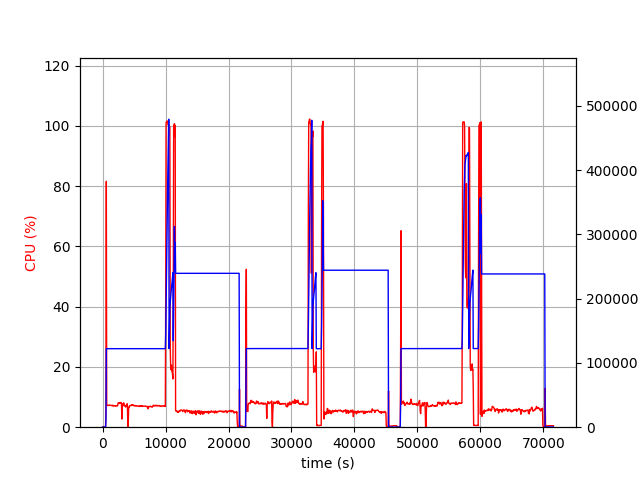}}     
    \caption{Memory usage during streaming of a 128GB large model. \label{fig:streaming_memory}}
\end{figure*}

In the following, we are using two clients: \textit{Site-1} with a fast connection and \textit{Site-2} with a slower one\footnote{The server was deployed on Azure with ~900GB RAM. Each client had $\sim$380GB RAM.}. Figure~\ref{fig:streaming_memory} illustrates the server's and clients' memory usage during the FL training run over three rounds.

The local training task was to add a small number to those arrays. The aggregator on the server side was not changed. This job required at least two clients and ran three rounds to finish. During the experiment, the server used over 512GB, i.e., 128GB $\times$ 2 (clients) $\times$ 2 (model and runtime space).  Although most of the time, the server was using less than 512GB, there were a few peaks that reached 700GB or more.

The \textit{Site-1} client, with its fast bandwidth connection with the server, received and sent the models in about 100 minutes and entered a nearly idle state with little CPU and memory usage after the communication ended. Both clients used about 256GB, i.e., $128GB \times 2$ (for model and runtime space), but at the end of receiving large models and at the beginning of sending large models, these two clients required more than 378GB, i.e., 128GB $\times$ 3.

\subsection{Federated PEFT Performance}
\begin{figure}[htbp]
    \centering
     \subfloat[\centering $\alpha=1.0$]{\includegraphics[width=0.33\textwidth,valign=m]{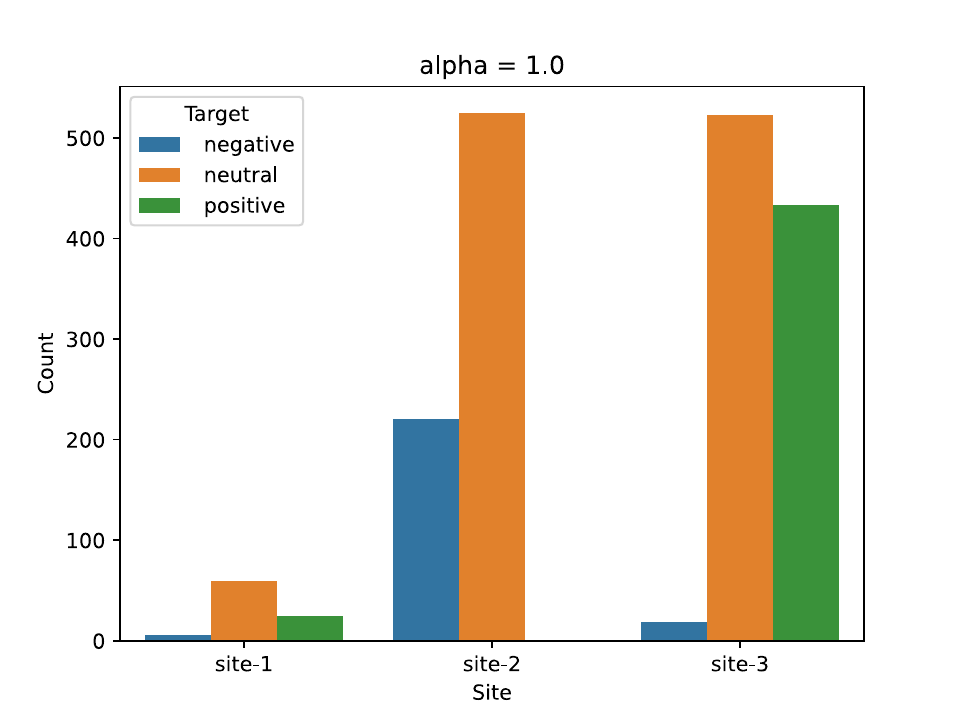}}
     \hfill
     \subfloat[\centering $\alpha=5.0$]{\includegraphics[width=0.33\textwidth,valign=m]{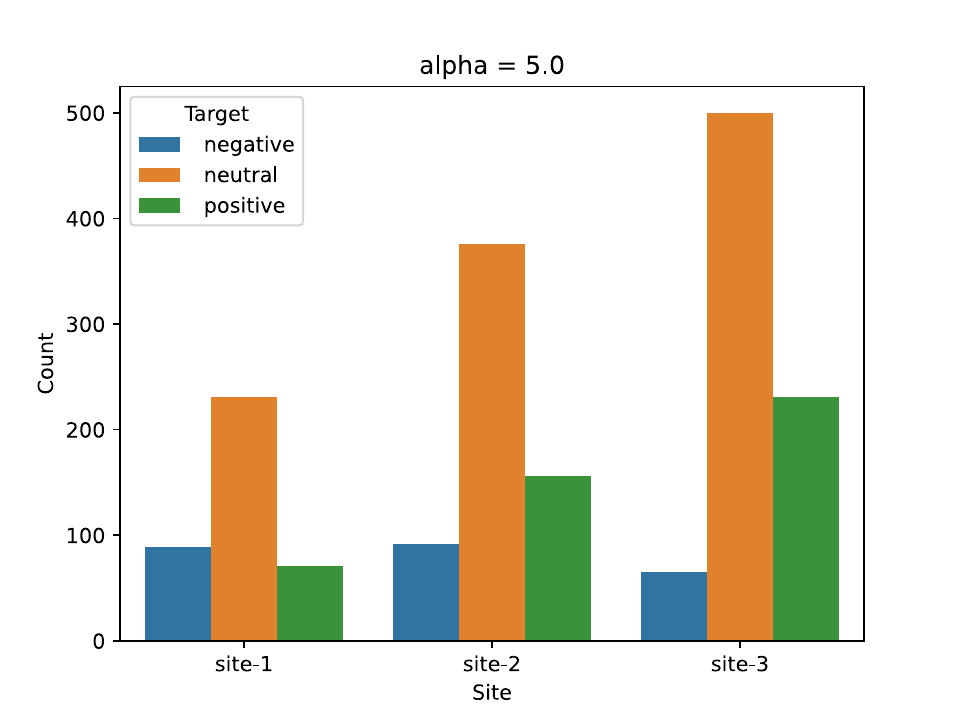}}
     \hfill
     \subfloat[\centering $\alpha=10.0$]{\includegraphics[width=0.33\textwidth,valign=m]{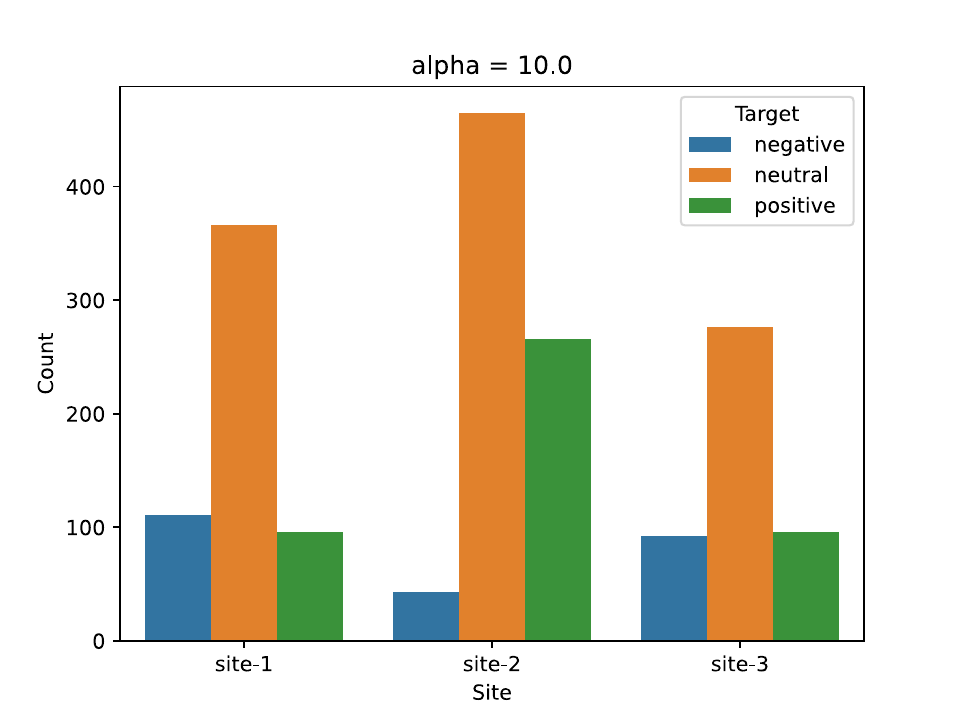}}     
    \caption{Simulation of different data distributions among clients. \label{fig:peft_alpha}}
\end{figure}

For PEFT, we utilize NeMo's PEFT methods. With a single line of configuration change, you can experiment with various PEFT techniques, such as p-tuning~\cite{liu2023gpt}, adapters~\cite{houlsby2019parameter}, or Low-Rank Adaptation (LoRA)~\cite{hu2021lora}, all of which introduce a small number of trainable parameters to the LLM. These parameters condition the model to generate the desired output for the downstream task. These approaches minimize resource utilization by training a smaller subset of parameters compared to full-finetuning. We train a GPT Megatron model with 345 million parameters on a financial sentiment prediction task~\cite{malo2014good} using LoRA. In total, this data contains 1,800 pairs of headlines and corresponding sentiment labels. We use a Dirichlet sampling strategy~\cite{wang2020federated} for creating a heterogeneous data partition among the clients. Examples of how the training data is distributed among the three clients using different values of $\alpha$ are shown in Fig.~\ref{fig:peft_alpha}.

\begin{figure}[htbp]
    \centering
     \subfloat[\centering $\alpha=1.0$]{\includegraphics[width=0.33\textwidth,valign=m]{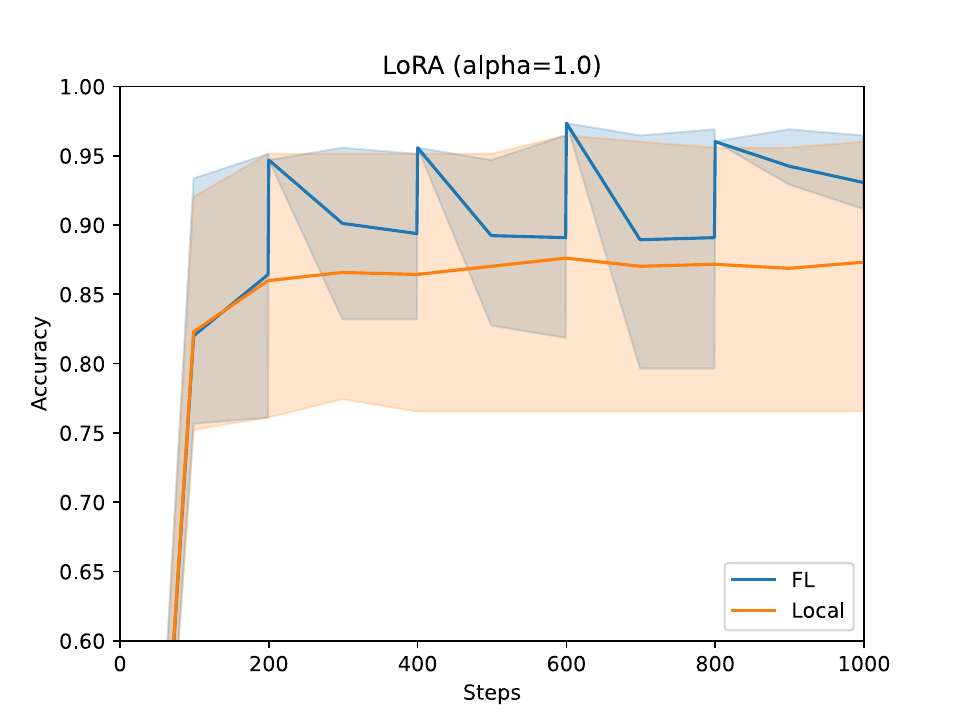}}
     \hfill
     \subfloat[\centering $\alpha=5.0$]{\includegraphics[width=0.33\textwidth,valign=m]{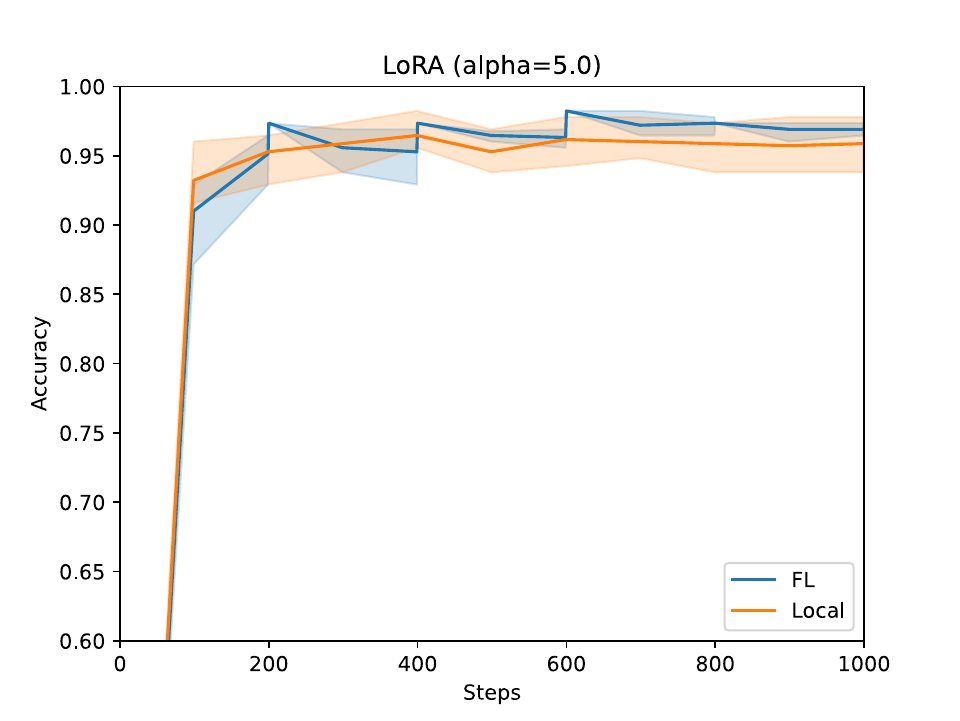}}
     \hfill
     \subfloat[\centering $\alpha=10.0$]{\includegraphics[width=0.33\textwidth,valign=m]{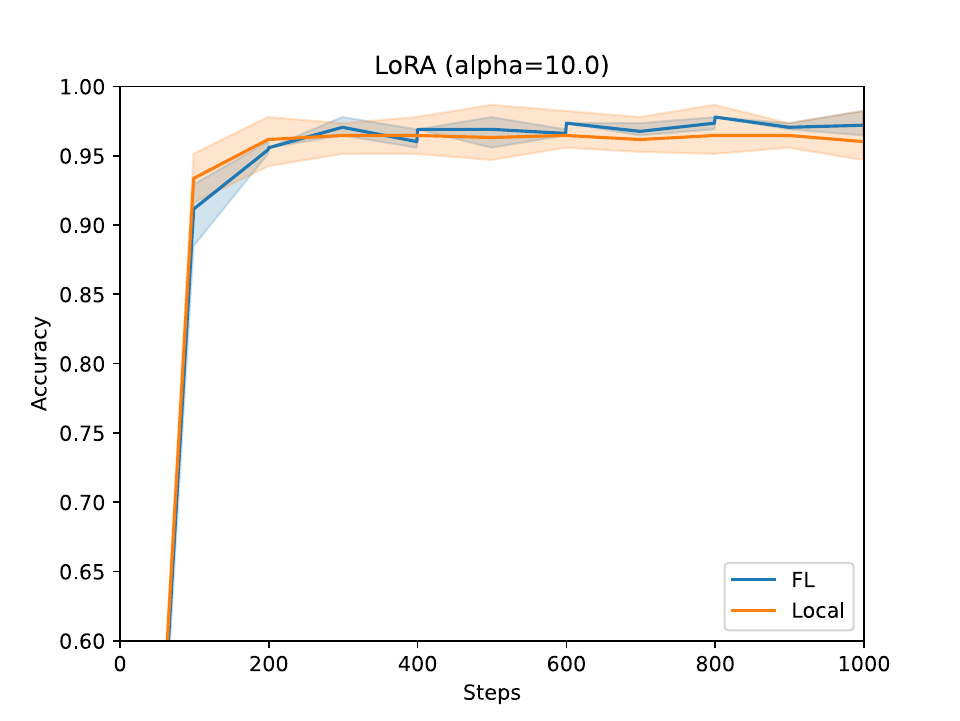}}     
    \caption{PEFT accuracy curves on clients using their ``Local'' data alone versus the accuracy when training a joint model using “FL” that can learn from the data available at all sites without having to centralize the data.  \label{fig:peft_accuracy}}
\end{figure}

Figure~\ref{fig:peft_accuracy} shows examples of how the training data is distributed among the three clients when using different $\alpha$ values. The lines show the mean accuracy of local models during training, and shaded areas indicate the 95\% confidence interval.

Example input headlines from the financial sentiment prediction tasks and the predictions from the trained global model are shown in \texttt{\textbf{bold}}:

\begin{itemize}
\small
    \item \texttt{The products have a low salt and fat content.\\ \textbf{sentiment: neutral}}
    \item \texttt{The agreement is valid for four years.\\ \textbf{sentiment: neutral}}
    \item \texttt{Diluted EPS rose to EUR3 .68 from EUR0 .50.\\ \textbf{sentiment: positive}}
    \item \texttt{The company is well positioned in Brazil and Uruguay.\\ \textbf{sentiment: positive}}
    \item \texttt{Profit before taxes decreased by 9\% to EUR 187.8 mn in the first nine months of 2008, compared to EUR 207.1 mn a year earlier.\\ \textbf{sentiment: negative}}
\end{itemize}

\subsection{Federated SFT Performance}
 
\begin{figure}[htbp]
    \centering
    \includegraphics[width=0.95\columnwidth]{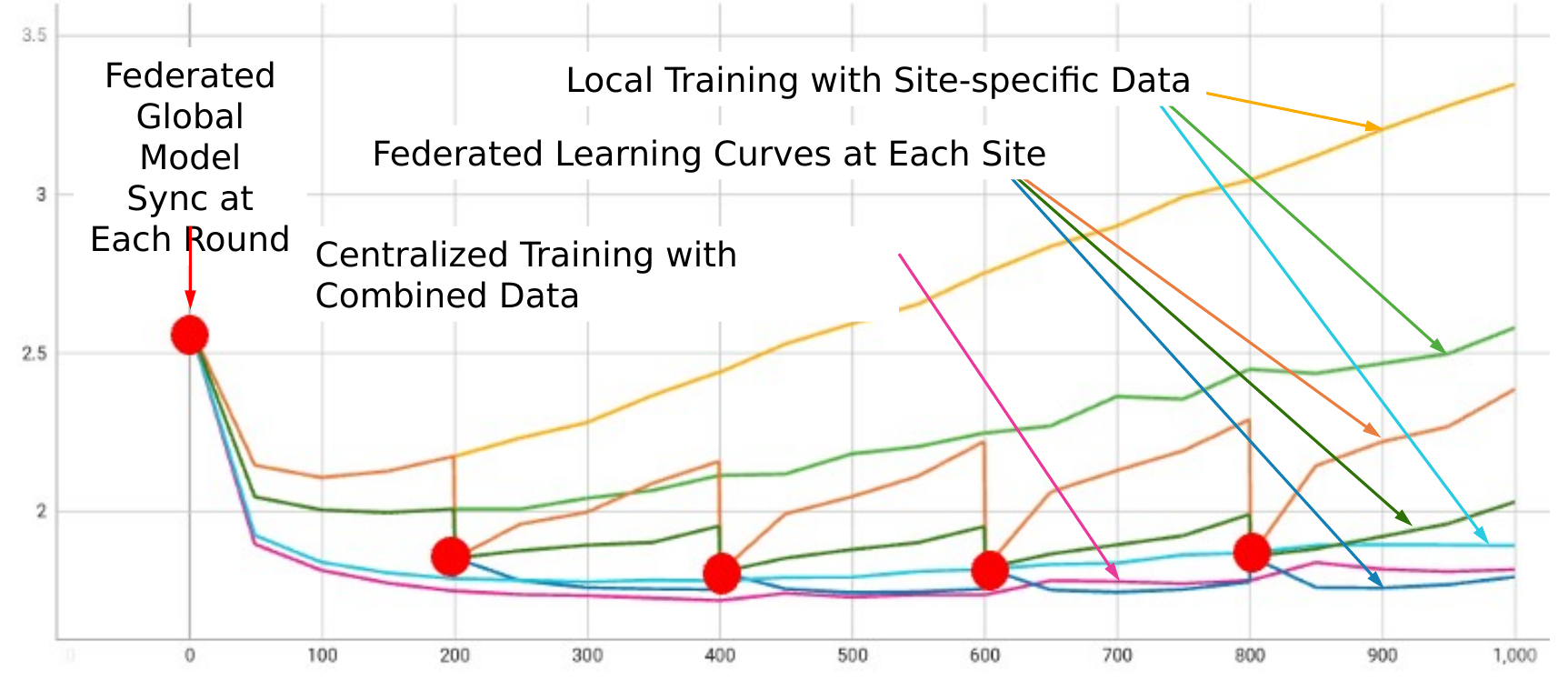}
    \caption{SFT validation loss curve. \label{fig:sft}}
\end{figure}

For SFT, we conducted experiments using the \textit{nemo-megatron-gpt-1.3B}\footnote{\url{https://huggingface.co/nvidia/nemo-megatron-gpt-1.3B}} model, SFT for five rounds, training on three open datasets Alpaca\footnote{\url{https://huggingface.co/datasets/tatsu-lab/alpaca}}~\cite{alpaca}, databricks-dolly-15k\footnote{\url{https://huggingface.co/datasets/databricks/databricks-dolly-15k}}~\cite{DatabricksBlog2023DollyV2}, OpenAssistant Conversations\footnote{\url{https://huggingface.co/datasets/OpenAssistant/oasst1}}~\cite{kopf2023openassistant}, one for each client.

Figure~\ref{fig:sft} illustrates the validation curves under all experiment settings: local-only training on each of the three datasets, on a combined dataset, and federated learning with all three clients training together using the FedAvg algorithm. Smooth curves represent local training, while ``step curves'', identified by red dots, are for FL - the ``steps'' are due to global model aggregation and update at the beginning of each FL round.

Evaluating LLMs can be a non-trivial task. Following popular benchmark tasks, we perform three language modeling tasks under zero-shot settings, including HellaSwag (H)~\cite{zellers2019hellaswag}, PIQA (P)~\cite{bisk2020piqa}, and WinoGrande (W)~\cite{sakaguchi2021winogrande}. Table~\ref{tab:sft} shows the results of each SFT model, with ``BaseModel'' representing the model before SFT. 
We utilize both ``unnormalized'' ($*_\mathrm{acc}$) and ``normalized'' ($*_\mathrm{\widehat{acc}}$) metrics and compute their mean for overall evaluation~\cite{eval-harness}.
As shown, FL can help achieve the best overall performance compared to the models fine-tuned on the individual datasets by effectively combining updates from diverse sources without having to centralize the data as in the ``Combined'' setting.

\definecolor{Shark}{rgb}{0.121,0.137,0.156}
\begin{table}
\caption{Model performance on three benchmark tasks: HellaSwag (H), PIQA (P), and WinoGrande (W). \label{tab:sft}}
\centering
\small
\begin{tabular}{|l|c|c|c|c|c|c|}
\toprule
          & $H_\mathrm{acc}$ & $H_\mathrm{\widehat{acc}}$ & $P_\mathrm{acc}$ & $P_\mathrm{\widehat{acc}}$ & $W_\mathrm{acc}$ & Mean \\
\midrule
BaseModel & 0.357  & 0.439      & 0.683       & 0.689     & 0.537      & 0.541         \\
\midrule
Alpaca    & 0.372  & 0.451      & 0.675       & 0.687     & 0.550      & 0.547         \\
Dolly     & 0.376  & \textbf{0.474} & 0.671   & 0.667     & 0.529      & 0.543         \\
Oasst1    & 0.370  & 0.452      & 0.657       & 0.655     & 0.506      & 0.528         \\
\midrule
Combined  & 0.370  & 0.453      & 0.685       & \textbf{0.690}     & 0.548      & 0.549         \\
FedAvg    & \textbf{0.377}  & 0.469      & \textbf{0.688}       & 0.687     & \textbf{0.560}      & \textbf{0.556}         \\
\bottomrule
\end{tabular}
\end{table}

\subsection{Subcellular Structure Prediction}
\label{sec:scl_results}
First, we run federated inference of the ESM-1nv model to extract embeddings, which requires a GPU with at least 12GB memory. 

\begin{figure}[htbp]
    \centering
    \includegraphics[width=0.99\columnwidth]{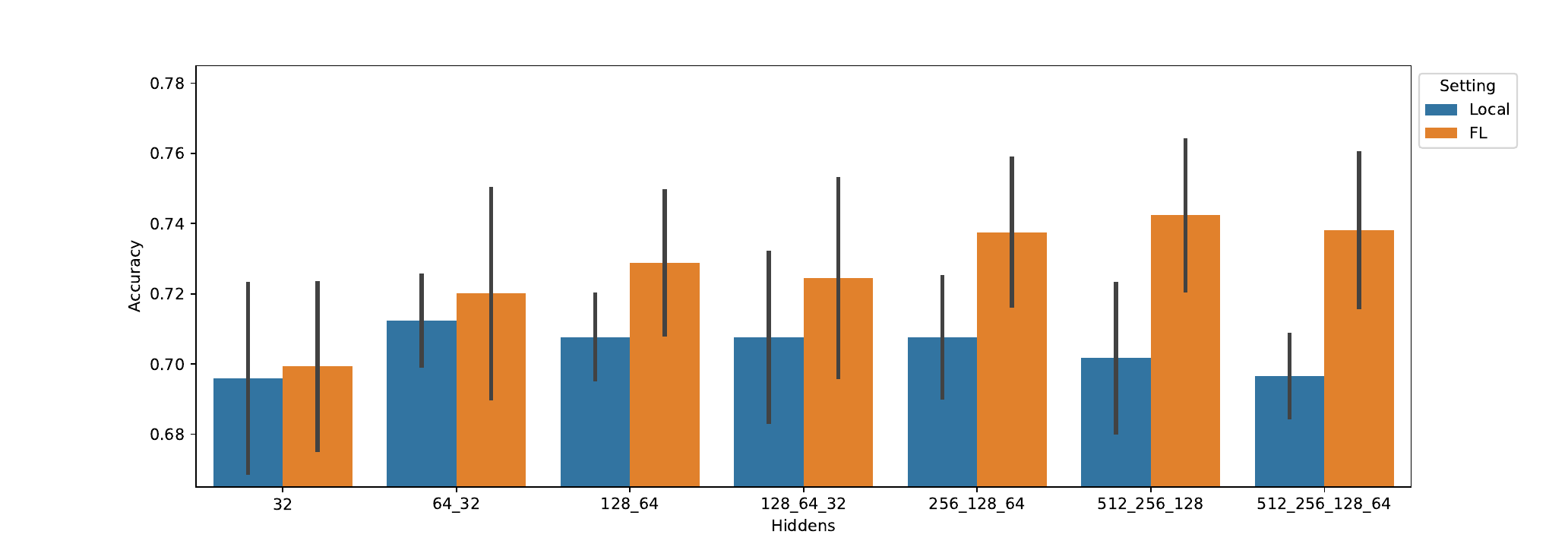}
    \caption{Subcellular Structure Prediction of local and global models (using FL). The bar plots show the mean and standard deviation error of the accuracy across clients. \label{fig:scl_results}}
\end{figure}

Next, we want to classify proteins for their subcellular location. Hence, we train a simple \textit{scikit-learn}~\cite{scikit-learn} Multi-layer Perceptron (MPL) classifier on top of the extracted ESM-1nv features using FedAvg. The MLP model uses a network of hidden layers to fit the input embedding vectors to the model classes (the cellular locations above). The results in Fig.~\ref{fig:scl_results} show the local (clients train on their local data alone) and global (using FedAvg) model performances varying the number of 32 hidden units of the MLP from one layer with 32 hidden units to four layers containing 512, 256, 128, and 64 units, respectively.

As the MLP parameters increase, the local models tend to overfit to the training data, while the FL models can benefit from the larger effective training set sizes and perform well on the test sets. 


\section{Conclusion}
FL presents exciting opportunities for customizing foundational LLMs and tackling data challenges while prioritizing privacy. Fine-tuning techniques, which aim to adapt foundational LLMs for specific domains and tasks, can be seamlessly applied within an FL paradigm, leveraging the broader availability of diverse distributed datasets. NVFlare offers communication support to facilitate collaborative LLM training. These techniques, when combined with advancements in model development, pave the way for more adaptable and efficient LLMs.

This paper primarily highlights NVFlare's new features, demonstrating its capability to simplify and scale the adaptation of LLMs with popular fine-tuning approaches like PEFT and SFT within FL, thereby supporting federated training of massive models. Two key features stand out: the \textit{Client API} and the ability to stream large datasets. 

Utilizing the \textit{Client API}, it becomes straightforward to translate existing code developed for centralized training into a federated scenario. It avoids the need for a major restructuring of the training code to fit certain client class structures as required by other FL frameworks.

The \textit{Streaming API} enables the communication of arbitrary large message sizes, paving the way for enabling FL for massive models, such as modern LLMs.

FL holds promise for collaborative learning, preserving privacy, and enhancing model performance. With NVFlare, federated workflows can be seamlessly transitioned into real-world production environments. For further details, please visit the code repository at \url{https://github.com/NVIDIA/NVFlare}.

\bibliographystyle{abbrv}
\bibliography{ref}

\end{document}